\documentclass[sigconf, nonacm]{acmart}
\setcopyright{none}
\renewcommand\footnotetextcopyrightpermission[1]{}

\usepackage{booktabs}
\usepackage{tabularx}
\usepackage{array}
\usepackage{svg}
\usepackage{float}

\newcolumntype{Y}{>{\raggedright\arraybackslash}X}

\AtBeginDocument{%
  \providecommand\BibTeX{{%
    \normalfont B\kern-0.5em{\scshape i\kern-0.25em b}\kern-0.8em\TeX}}}

\graphicspath{{./images/}}

\renewcommand\footnotetextcopyrightpermission[1]{}

\begin{document}

\title[Adapting Prithvi-EO for Fallow Detection]{Adapting Prithvi-EO for Fallow Detection for Food-Water Nexus: ViT-Adapter Necks and Parameter-Efficient Backbone tuning of Geospatial Foundation Model
}


\author{Sk Muhammad Asif}
\affiliation{%
  \institution{Earth, Atmospheric and Geospatial Science}
  \institution{Saint Louis University}
  \city{Saint Louis}
  \state{Missouri}
  \country{USA}
  \postcode{63103}}
\email{skmuhammad.asif@slu.edu}

\author{Orhun Aydin}
\affiliation{%
  \institution{Earth, Atmospheric and Geospatial Science}
  \institution{Saint Louis University}
  \city{Saint Louis}
  \state{Missouri}
  \country{USA}
  \postcode{63103}}
\email{orhun.aydin@slu.edu}

\renewcommand{\shortauthors}{Asif and Aydin}

\begin{abstract}
  Understanding spatial distribution of fallow land is important for optimizing the food-water (FW) nexus, given fallowing’s role in crop rotation and water conservation. Fallow is a low accuracy class in USDA Cropland Data Layer (CDL).  Geospatial foundation model (GFM), Prithvi-EO has shown strong transferability across computer vision tasks. However, its Vision Transformer (ViT) backbone produces features at a single spatial scale that are ill-suited for the multi-scale features required by object detection heads. Existing approaches synthesise multi-scale pyramids through scaling of single-stride tokens, sacrificing spatial heterogeneity, and full backbone fine-tuning is computationally prohibitive for GFMs. We evaluate a fallow detection pipeline combining two parameter-efficient fine-tuning (PEFT) schemes: Low-Rank Adaptation (LoRA) and a hybrid PEFT, with three neck designs: pseudo multi-scale, Lite ViT-Adapter, and Full ViT-Adapter. Our best configuration, Lite ViT-Adapter with a one-stage head, achieves a mAP@50 of 0.9479 with the Diou loss, suggesting the effectiveness of center-aware localization for irregular fallow field detection. ViT-Adapter free one-stage detection under LoRA improves the adapter-free anchor-based approach by 6.42\%, and the best configuration improves baseline adapter-free anchor-based approach by 25.70\%. These results demonstrate that lightweight spatial prior fusion and selective backbone unfreezing enable Prithvi-EO to capture local fallow patterns more effectively, outperforming approaches that rely on reshaped single-stride ViT tokens.

\end{abstract}



\maketitle

\section{Introduction}

Food–water (FW) nexus is a burgeoning research area that addresses the emerging trade-offs between food and water resources, thereby providing a scientific basis for integrated, cross-sectoral policy making ~\cite{albrecht2018water}. Fallowing, intentional extension of non-cropped periods of crop lands~\cite{greb1979reducing}, is deemed as a critical lever in boosting agricultural productivity~\cite{zarczynski2023role}, and for centuries it has been an integral part of agriculture~\cite{white1970fallowing}. Fallowing further serves as a water conservation measure, operating directly 
through soil moisture retention~\cite{zeleke2017fallow} and indirectly through reduced irrigation pressure in water-stressed regions~\cite{song2023cropland}. Due to fallowing's dual role in the intersection of two important resources, namely food and water, it has been integrated into policies for sustainable food-water (FW) nexus management~\cite{jiang2024trade}. Internationally, as part of the National Food Security Mission (NFSM) by the United Nations, rice fallowing is proposed in regions where post-monsoon cultivation is limited due to water scarcity~\cite{jain2023water}. Locally, fallowing is incorporated into policy in the USA, as observed in the Upper Colorado River Basin, where water allocations are adjusted to address emerging water scarcity for farming communities~\cite{hung2022investigating}. At the policy level, large-scale monitoring of fallow extent is therefore a prerequisite for equitable water-sharing agreements and reliable assessment of multi-cropping patterns while simultaneously supporting sustainable agricultural productivity and freshwater availability~\cite{siebet2010global}.

Despite its agronomic and hydrological importance, fallow land is difficult to monitor accurately at scale~\cite{tong2020forgotten}. We encounter three fundamental challenges in the literature. Firstly, unlike various crop types, fallow fields do not have a specific spectral signature, and domain-adaptation problems with them are more pronounced compared to other crops~\cite{peng2022domain}. Fallow fields exhibit highly variable spectral signatures depending on their age, environmental conditions, and management practices~\cite{begue2018remote} thereby, historically remaining an ambiguous category in the official USDA CDL~\cite{lark2017measuring}. Variations in fallow terminology or nomenclature ambiguity is another issue that further complicates the challenges related to fallow mapping~\cite{subedi2022drivers}. Daskalova et al.~\cite{daskalova2023abandoning} mention the absence of concrete definitions in land abandonment practices and distinguish land abandonment from fallow lands with a temporal threshold of 5 years. Conversely, some studies identify long abandonment as long-term fallow~\cite{kozak2021impact, salamon2011plant}. Thirdly, temporal variability: a parcel may cycle in and out of fallow status within a single growing season, limiting the utility of single-date observations~\cite{yin2020monitoring}. Existing approaches address some of these challenges through supervised classification of multispectral time series~\cite{tong2020forgotten}, phenological compositing~\cite{foerster2012crop}, or multi-temporal segmentation~\cite{tong2020forgotten}. However, these pixel- or segmentation-oriented approaches primarily emphasize class assignment and boundary delineation, while less directly capturing fallow fields as discrete spatial units; prior Sentinel-2 cropland-mapping work has shown that object-based representations can be more computationally efficient and better suited for field-level vector outputs~\cite{belgiu2018sentinel}.

Geospatial foundation models (GFMs), large-scale vision transformers pre-trained on satellite image archives~\cite{vatsavai2024geospatial} offer a promising route to more robust fallow detection. Because GFMs encode spectral and phenological patterns from globally diverse scenes~\cite{hong2024spectralgpt}, they are better positioned than task-specific models to handle the spectral ambiguity and geographic variability  that confound fallow classification~\cite{ma2026harvesting}. Yet the application of GFMs to fallow land detection, localising and bounding individual parcels, remains largely unexplored. The shapes and size of fallow lands vary widely across geographic regions~\cite{oliphant2024automated} and time~\cite{song2022mapping}. Additionally, fallow is treated as a part of crop rotation in different crop systems across USA ~\cite{anderson1999alternative}. Object detection is a natural fit for policy-relevant fallow monitoring: it yields spatially grounded parcel-level summaries~\cite{zheng2026comprehensive}, and directly supports field-scale crop-rotation inference~\cite{upcott2023new} and irrigation-allocation estimates~\cite{chen2023mapping}.

In this paper, we present an end-to-end fallow detection pipeline built on Prithvi-EO-2.0~\cite{szwarcman2025prithvieo2}, a GFM pre-trained on Harmonized Landsat Sentinel-2 (HLS) imagery. Using a single HLS tile over eastern Colorado (T13TGE, 2022) with USDA Cropland Data Layer (CDL) ground truth~\cite{boryan2011monitoring}, we systematically evaluate how backbone adaptation strategy, multi-scale neck design, and detection head choice affect detection accuracy under realistic computational constraints. Our principal contributions are:
\begin{itemize}
  \item  End-to-end fallow detection pipeline over six-band HLS Sentinel-2 imagery, combining CDL-derived bounding-box annotations with a Prithvi-EO-2.0 backbone.
  \item A systematic comparison of two parameter-efficient adaptation schemes (LoRA-only and Hybrid PEFT) across three multi-scale neck architectures (pseudo multi-scale, Lite ViT-Adapter, Full ViT-Adapter) and two detection heads (Faster R-CNN and fully convolutional one stage (FCOS)).
  \item Identification of the accuracy–efficiency trade-off and proposing the best configuration under computational bottleneck. 
\end{itemize}

\section{Related Work}
Large-scale self-supervised pretraining on satellite imagery has established GFMs as the dominant paradigm for Earth observation. Jakubik et al. \cite{jakubik2023foundation} introduced Prithvi-EO-1.0, a masked-autoencoder ViT pre-trained on over 1 TB of HLS imagery across six spectral bands, demonstrating strong data efficiency on flood mapping, wildfire segmentation, and  crop classification. Szwarcman et al. \cite{szwarcman2025prithvieo2} scaled this to Prithvi-EO-2.0 (300M parameters), expanding the training corpus to 4.2 million global HLS time-series samples and introducing temporal and geographic location embeddings, yielding an 8\% improvement over its predecessor across tasks at multiple spatial resolutions. Nguyen et al.~\cite{nguyen2025foundation} utilized four foundation models including Prithvi-EO-2.0 for cloud masking evaluating their impact of cloud segmentation. Gao et al.~\cite{gao2026introduction} compiled contributions dedicated towards advancing the state of the art GFMs and their applications.

Plain ViT backbones generate token embeddings at a single spatial scale, mismatching object detection heads that require multi-scale feature pyramids~\cite{szwarcman2025prithvieo2}. Hsu et al.~\cite{hsu2025geospatial} addressed this for Prithvi-EO-1.0 by constructing a multi-scale feature generation module, then evaluating a Mask R-CNN-based pipeline across four remote sensing benchmarks including EuroCrops~\cite{schneider2023eurocrops}. Their best results used full backbone fine-tuning; the present study departs from this by exploring parameter-efficient adaptation to reduce the number of trainable parameters.

Hu et al.~\cite{hu2022lora} proposed LoRA, which injects trainable low-rank decomposition matrices into frozen transformer weight projections, matching full fine-tuning performance with orders-of-magnitude fewer trainable parameters. Marti-Escofet et al.~\cite{marti2025fine} conducted the first systematic evaluation of parameter-efficient fine-tuning (PEFT) methods on GFMs across five Earth observation   datasets, finding that LoRA and ViT-Adapter-based techniques match or exceed full fine-tuning on in-distribution data while improving generalisation to geographically unseen  regions. 

Plain ViTs lack local spatial inductive biases critical for dense prediction. Chen et al.~\cite{chen2022vision} proposed ViT-Adapter, which augments a frozen ViT with a convolutional spatial
prior module and bidirectional injector-extractor cross-attention at multiple backbone depths, achieving 60.9 box AP on COCO without additional detection data. 

Ren et al.~\cite{ren2015fasterrcnn} introduced Faster R-CNN, in which an RPN generates candidate regions refined by per-proposal classification and regression heads; paired with Feature Pyramid Network (FPN), it
remains a standard benchmark in remote sensing detection. Tian et al.~\cite{tian2019fcos} proposed FCOS, an anchor-free detector that predicts class scores, box offsets, and centerness densely at each FPN pyramid level, eliminating anchor design and the proposal stage. Generalised IoU (GIoU)~\cite {rezatofighi2019generalized} and Distance IoU (DIoU)~\cite{zheng2020distance} box regression losses are two widely adopted loos functions in object detection workflows. XiaoFan et al.~\cite{xiaofan2019introduce} adopted GIoU to improve the location of predicted bounding box. Cao~\cite{cao2021experimental} evaluated the effect of GIoU and DIoU losses for different object detection algorithms.

\section{Methodology}

\subsection{System overview}
\begin{figure}[h!]
    \centering
    \includegraphics[width=0.5\textwidth]{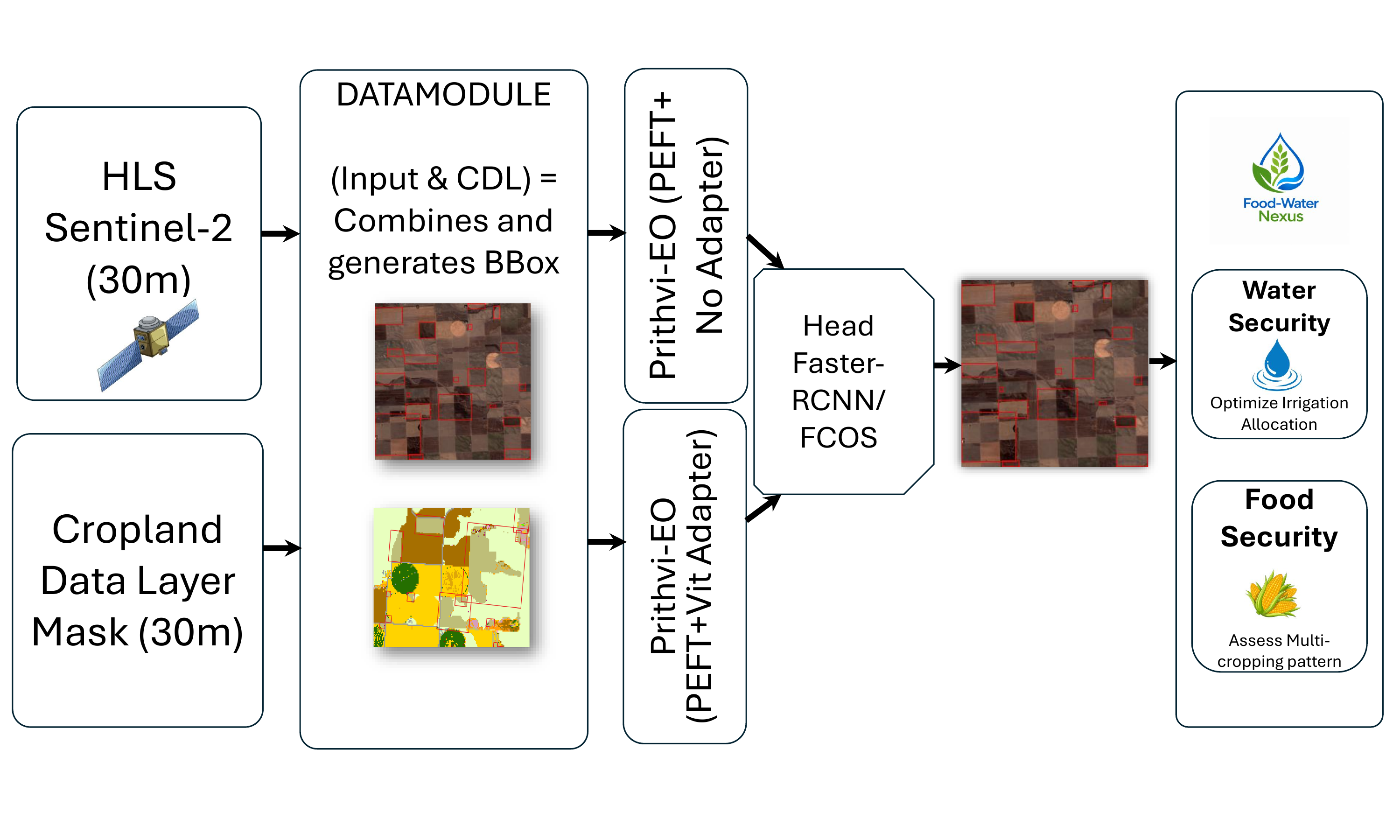}
    \caption{End-to-end fallow land detection pipeline. Data sources → label generation → backbone adaptation → detection heads → predictions → food–water nexus applications.}
   
    \label{fig:pipeline}
\end{figure}

We design an end-to-end fallow detection pipeline built on Prithvi-EO-2.0, a geospatial foundation model (GFM) pretrained on six-band Harmonized Landsat and Sentinel-2 (HLS) imagery, to support food-water nexus optimization through spatially explicit mapping of fallow agricultural land. As depicted in Figure~\ref{fig:pipeline}, the pipeline consists of three stages: a data module that converts USDA Cropland Data Layer (CDL) masks into bounding box labels; a backbone stage that applies one of two parameter-efficient fine-tuning (PEFT) schemes to the Prithvi-EO-2.0 encoder while introducing ViT-Adapter; and a detection head (either FCOS or Faster-RCNN) that operates over multi-scale feature maps. All pipeline variants share the same data pipeline and six-band HLS input: B02, B03, B04, B8A, B11, and B12. 

\subsection{Data and label generation }
\label{sec:bbox_generation}
\begin{figure}[h!]
    \centering
    \includegraphics[width=0.5\textwidth]{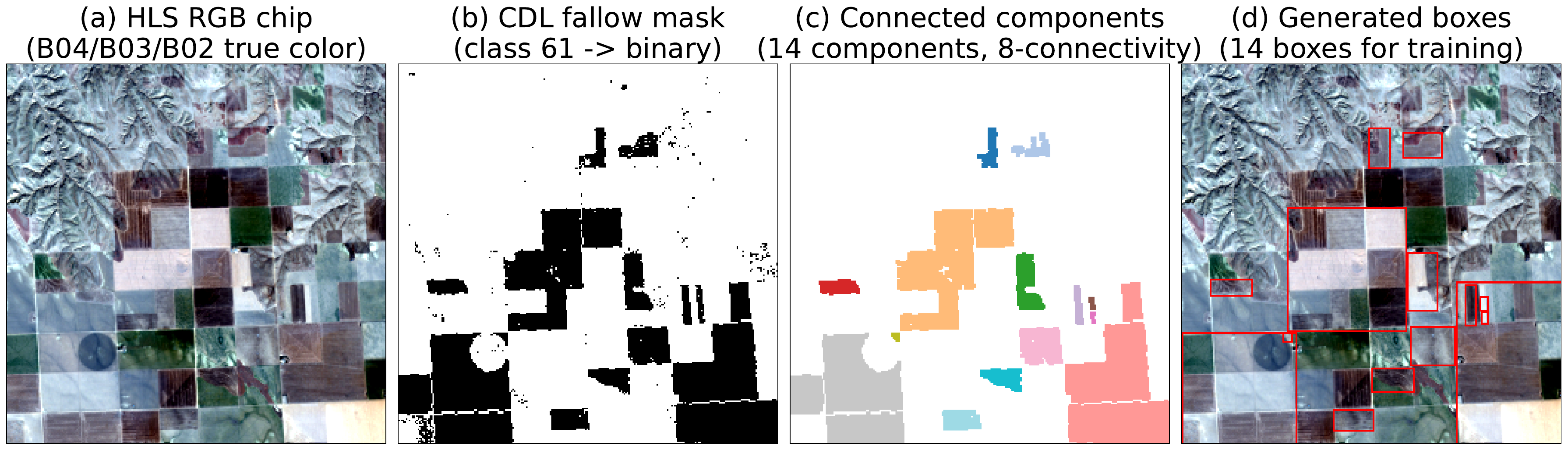}
    \caption{CDL-based bounding-box generation. Fallow pixels are extracted from the CDL layer, grouped into connected parcels, and converted to bounding boxes.}
    \label{fig:bbox_generation}
\end{figure}

As no hand-labeled bounding box annotations exist for fallow parcels at the scale of this study, we derive ground truth automatically from the USDA Cropland Data Layer (CDL), an annually updated, 30 m resolution crop classification mask covering the contiguous United States. Our mask-to-bounding-box workflow proceeds in four steps. First, CDL pixels classified as fallow/idle cropland are extracted as a binary foreground mask. Second, a morphological opening (with a 3×3 square kernel) removes isolated noise pixels. Third, connected components are identified under 8-connectivity, grouping edge- and corner-adjacent fallow pixels into individual parcels. Fourth, each component's axis-aligned minimum bounding rectangle is recorded as a ground truth box. Components with fewer than 5 pixels or bounding boxes smaller than 25 
pixels² are discarded; no upper-size filter is applied, preserving large contiguous fallow blocks. Figure~\ref{fig:bbox_generation} illustrates the full pipeline from raw HLS chip to generated bounding boxes.

\subsection{Prithvi as detection backbone}
\label{sec:prithvi_as_backbone}
Prithvi-EO-2.0 was pretrained as a masked autoencoder on HLS time series, producing a sequence of patch token embeddings at a single spatial stride. This design is well-suited for classification and segmentation tasks but is misaligned with the requirements of object detection heads such as Faster R-CNN and FCOS, which expect multi-scale spatial feature maps at strides {4, 8, 16, 32}~\cite{lin2017feature}. Hsu et al.~\cite{hsu2025geospatial} previously adapted Prithvi-EO-1.0 for object detection, including on the EuroCrops agricultural benchmark, by reshaping intermediate token sequences into spatial maps and fine-tuning all modules end-to-end. We depart from this in two respects: we replace full fine-tuning with PEFT (section~\ref{sec:backbone_adaptation}), and we evaluate three neck designs to observe which design performs the best in addressing single-stride limitation  (section~\ref{multi-scale}).

\subsection{Multi-Scale Feature Construction}
\label{multi-scale}
To address the lack of true multi-scale feature discussed in section~\ref{sec:prithvi_as_backbone}, we evaluate three neck designs of increasing complexity: pseudo multi-scale (token reshaping and interpolation), Lite ViT-Adapter (output-level spatial fusion), and Full ViT-Adapter (bidirectional injector/extractor interaction). The pseudo multi-scale neck is used with Faster R-CNN and as the FCOS baseline; both ViT-Adapter variants are evaluated with FCOS only.

\subsubsection{Pseudo Multi-Scale}
\label{sec:pseudo_multi_scale}
\begin{figure}[h!]
    \centering
    \includegraphics[width=0.5\textwidth]{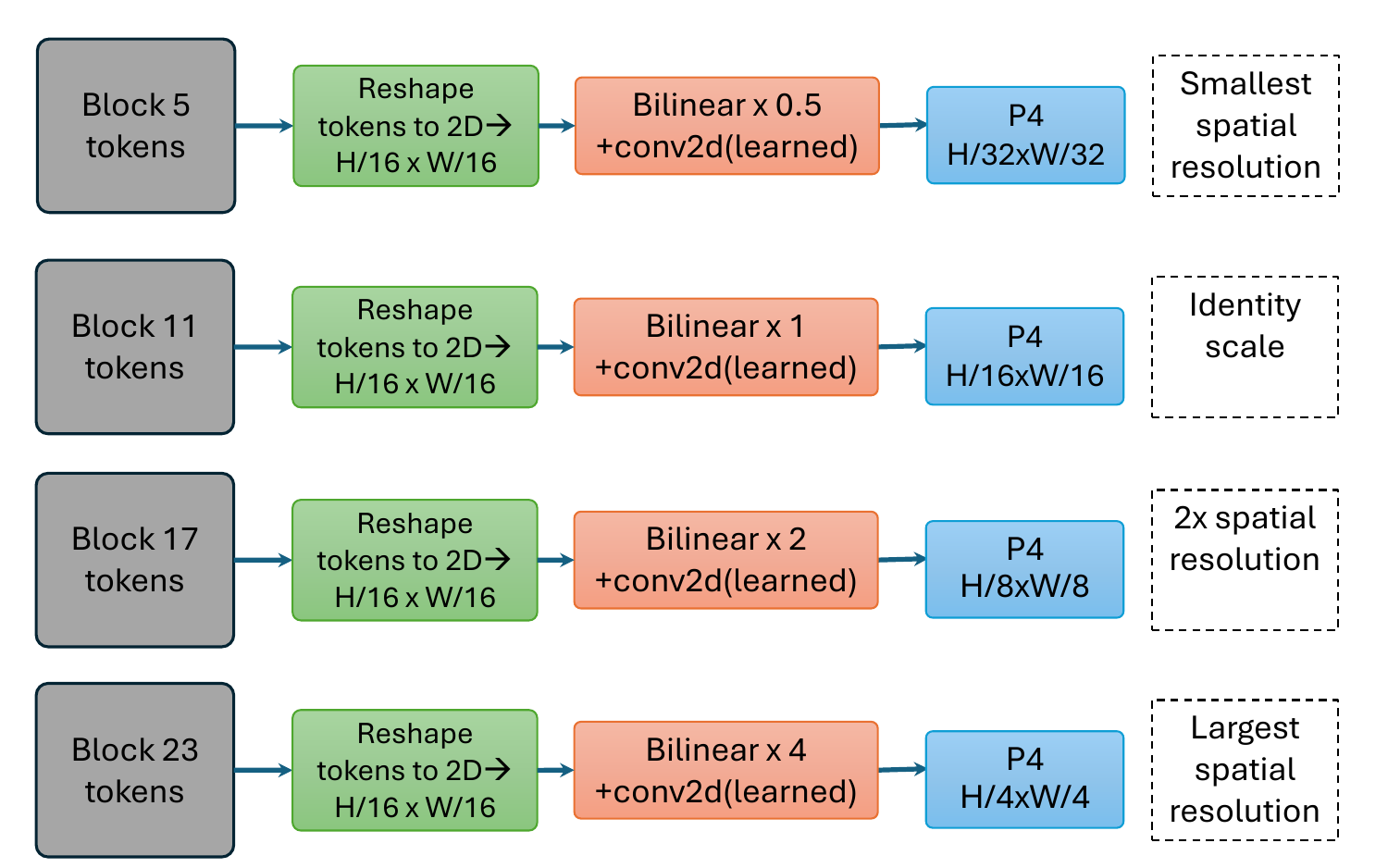}
    \caption{Pseudo multi-scale feature generation. Four token sequences taken from four ViT blocks, reshaped to H/16×W/16 and interpolated to four target resolutions.}
    \label{fig:token_reshape}
\end{figure}

We select intermediate token sequences from Prithvi-EO-2.0 blocks {5, 11, 17, 23}, reshape each from token-sequence to 2D spatial form, and pass them through a learned pyramidal interpolation module to produce four FPN-compatible feature maps at 256 channels. Because all four token sequences share the same spatial stride of 16 before interpolation, the resulting pyramid is synthesised rather than natively multi-scale — hence the term pseudo multi-scale. Figure~\ref{fig:token_reshape} illustrates the pipeline generating feature pyramids from single stride tokens of Prithvi-EO.

\subsubsection{Lite ViT-Adapter: Output-Only Additive Fusion}
Unlike the pseudo multi-scale neck, which generates FPN-compatible feature maps through a learned pyramidal interpolation module (section~\ref{sec:pseudo_multi_scale}), the Lite ViT-Adapter replaces that trainable component with a convolutional Spatial Prior Module (SPM) combined with parameter-free bilinear resize. The SPM is a fully trainable CNN of approximately 3.5 M parameters that runs in parallel with the Prithvi-EO-2.0 encoder, generating a hierarchy of convolutional feature maps $(c_1, c_2, c_3, c_4)$ at strides $\{4, 8, 16, 32\}$. The final Prithvi-EO-2.0 block's token embedding is bilinearly resized without any learned parameters to match each SPM scale and fused elementwise:
\begin{equation}
   \mathbf{F}_i = \mathbf{C}_i + \mathcal{R}_i(\mathbf{V}), \qquad i \in \{1,2,3,4\},
    \label{eq:lite_adapter_fusion}
\end{equation}

where $\mathbf{C}_i$ denotes the SPM feature map at scale $i$, $\mathbf{V}$ denotes the reshaped ViT feature map from the final Prithvi block, $\mathcal{R}_i(\cdot)$ denotes bilinear resizing to the spatial resolution of $\mathbf{C}_i$, and $\mathbf{F}_i$ is the fused feature map.

\begin{figure}[h!]
    \centering
    \includegraphics[width=0.5\textwidth]{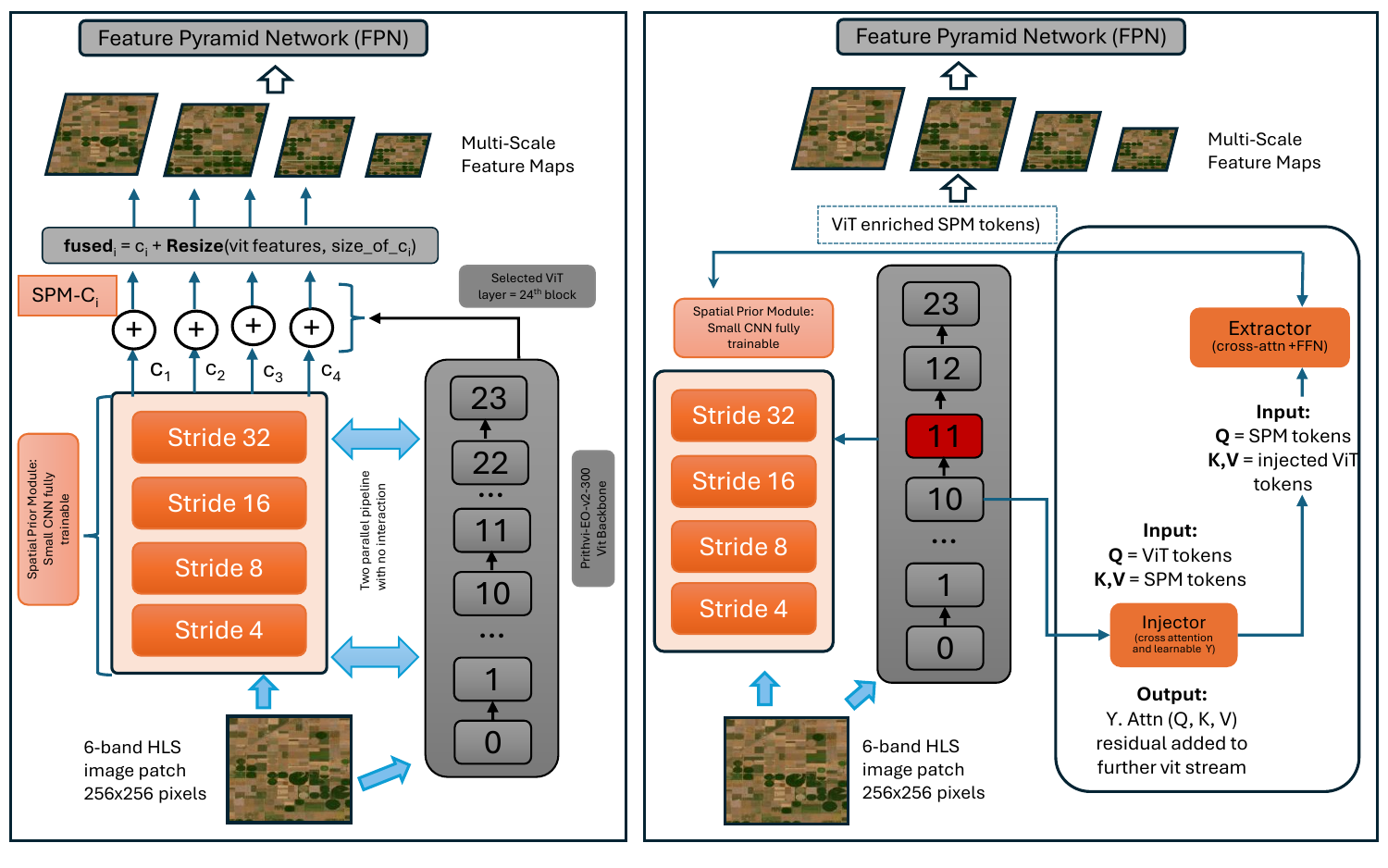}
    \caption{ViT-Adapter-based multi-scale feature generation. 
(\textit{Left}) Additive spatial fusion based Lite ViT-Adapter.
 (\textit{Right}) Bidirectional interaction based Full ViT-Adapter.}
    \label{fig:lite_vs_full_adapter}
\end{figure}

Finally, the four generated fused maps are passed directly to the FPN. No injector or extractor modules are introduced, keeping the adapter lightweight and straightforward to optimise. Figure~\ref{fig:lite_vs_full_adapter} demonstrates the operation for both Lite and Full ViT-Adapter.

\subsubsection{Full ViT-Adapter: Injector/Extractor Cross-Attention}

The Full ViT-Adapter replaces output-level fusion with repeated bidirectional interaction between the Prithvi token stream and the SPM at transformer block 11. Dedicated injector modules use cross-attention to incorporate SPM spatial priors into the ViT token stream; extractor modules use cross-attention in the reverse direction to update the SPM branch with intermediate transformer representations. This bidirectional coupling allows the final feature pyramid to carry both the global context learned during Prithvi-EO-2.0 pretraining and the local spatial structure captured by the convolutional branch. Relative to the Lite variant, the Full adapter is more expressive but incurs substantially higher engineering complexity, computational cost, and optimisation difficulty; block 11 is selected as the interaction point to balance semantic depth against computational budget.

\subsection{Backbone Adaptation Schemes}
\label{sec:backbone_adaptation}

We evaluate two PEFT schemes that control how much of the Prithvi-EO-2.0 encoder is updated during training. Both schemes keep the majority of backbone weights frozen and differ only in whether the final transformer blocks receive full gradient updates.

\subsubsection{LoRA-Only}
\label{sec:LoRA-Only}
Low-Rank Adaptation (LoRA)~\cite{marti2025fine} injects trainable rank-32 decomposition matrices into the query and value projections of every transformer block while keeping all original backbone weights frozen. For a projection weight $W \in \mathbb{R}^{1024 \times 1024}$, the adapted output is:
\begin{equation}
    h = W \cdot x + \left(\frac{\alpha}{r}\right) \cdot B \cdot A \cdot x
\end{equation}
where $A \in \mathbb{R}^{32 \times 1024}$ and $B \in \mathbb{R}^{1024 \times 32}$ are the trainable low-rank matrices and $\alpha/r = 1.0$. Applied across all 24 Prithvi blocks, LoRA introduces approximately 3.15~M backbone-trainable parameters while adding no inference latency~\cite{hu2022lora}.

\begin{figure}[h!]
    \centering
    \includegraphics[width=0.5\textwidth]{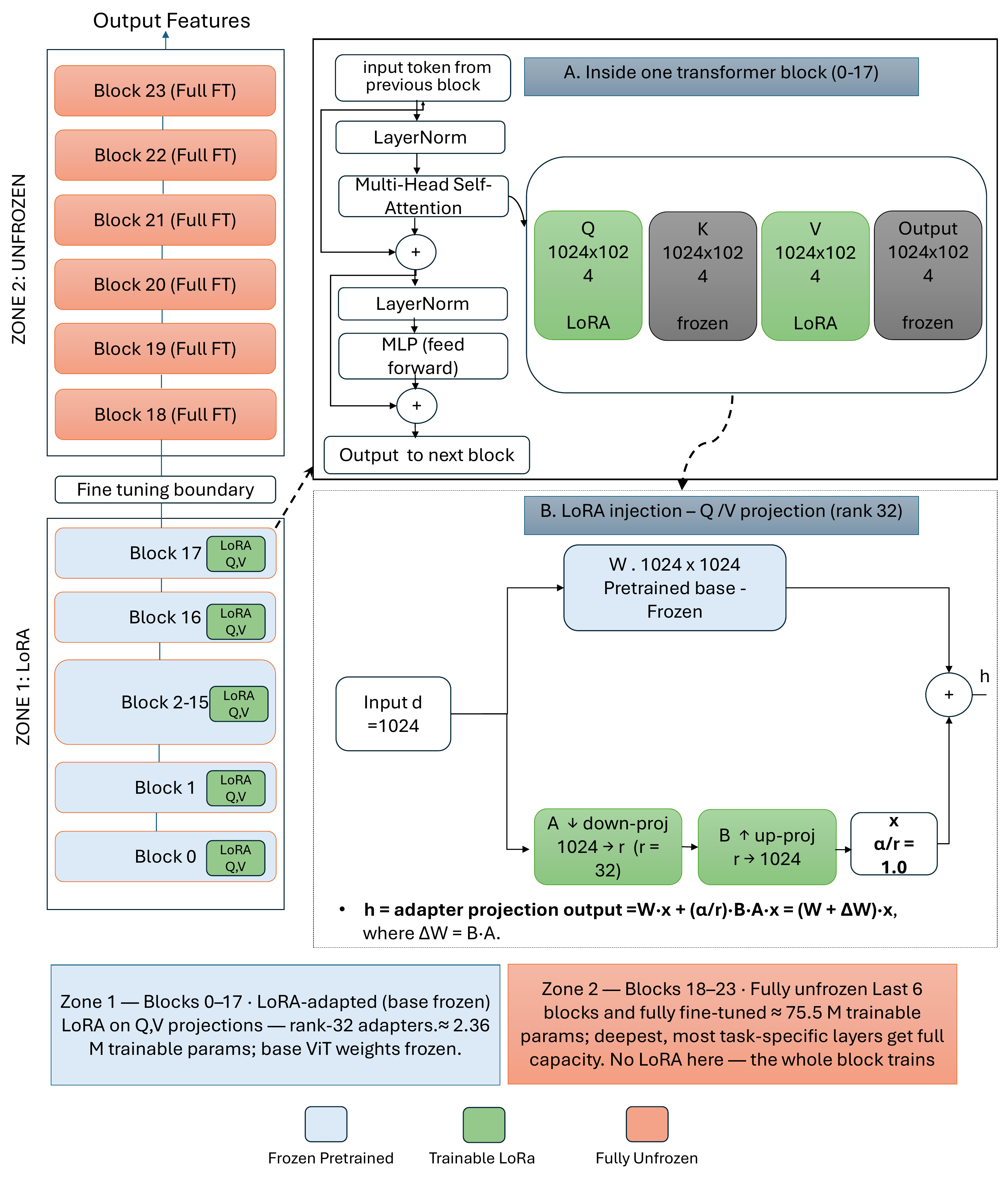}
    \caption{Hybrid PEFT adaptation strategy. Under Hybrid PEFT, blocks 0--17 applies the LoRA adapters while blocks 18--23 (Zone 2) are kept fully unfrozen.}
    \label{fig:hybrid_peft}
\end{figure}

\subsubsection{Hybrid-PEFT}
The Hybrid PEFT scheme splits the backbone into two zones. Zone 1 (blocks 0--17) retains LoRA-only adaptation as in section~\ref{sec:LoRA-Only}. Zone 2 (blocks 18--23) is fully unfrozen, allowing the deepest, most task-specific representations to adapt without low-rank constraints as shown in Figure~\ref{fig:hybrid_peft}. The motivation is   that FPN-based detection heads depend on semantically diverse feature maps across pyramid levels; fully unfreezing the final blocks allows higher-level Prithvi-EO-2.0 features to adapt more directly to fallow parcel geometry and texture. Table~\ref{tab:backbone_adaptation} summarises the parameter budget for LoRA-only scheme and for each zone of Hybrid-PEFT scheme.

\begin{table}[H]
\centering
\caption{Backbone adaptation strategy and trainable parameters.}
\label{tab:backbone_adaptation}
\small
\setlength{\tabcolsep}{5pt}
\renewcommand{\arraystretch}{1.3}
\begin{tabularx}{\columnwidth}{@{}l l X r@{}}
\toprule
\textbf{Zone} & \textbf{Blocks} & \textbf{Adaptation} & \textbf{Trainable} \\
\midrule
LoRA-only      & 0--23 & LoRA rank-32 on Q, V & $\approx$3.15 M \\
Hybrid Zone 1  & 0--17 & LoRA rank-32 on Q, V & $\approx$2.36 M \\
Hybrid Zone 2  & 18--23 & Full fine-tune       & $\approx$75.5 M \\
Hybrid Total   & 0--23  & Mixed                & $\approx$77.9 M \\
\bottomrule
\end{tabularx}
\end{table}

\subsection{Detection Heads}
\label{sec:detection_heads}
\subsubsection{Faster-RCNN}
We adopt Faster R-CNN~\cite{ren2015fasterrcnn} as the primary baseline which is a two-stage detector. The pseudo multi-scale neck (section~\ref{sec:pseudo_multi_scale}) produces four 256-channel FPN feature maps, over which a Region Proposal Network (RPN) generates candidate object regions. ROI-aligned features from each proposal are then classified and their bounding boxes regressed by dedicated detection heads. The detector is trained for a single foreground category namely the fallow field with background handled implicitly by the Faster R-CNN formulation. No ViT-Adapter variants are evaluated with this head to treat it as a traditional baseline approach and comparing the effects of modification with respect to this baseline.

\subsubsection{FCOS}
We adopt FCOS~\cite{tian2019fcos} as the anchor-free one-stage detector and the primary testbed for evaluating neck and tuning scheme choices. FCOS predicts class scores, bounding box offsets, and a centerness score densely at every spatial location across all FPN pyramid levels, eliminating  anchor design and the proposal generation stage entirely. Centerness down-weights off-center predictions, which is particularly beneficial for fallow parcels that are often irregular or elongated. The baseline FCOS configuration uses the pseudo multi-scale neck (section~\ref{sec:pseudo_multi_scale}); we additionally evaluate all three neck designs under both tuning schemes, yielding the five FCOS configurations summarised in section~\ref{sec:experimental_setup}.

\subsection{Experimental Setup}
\label{sec:experimental_setup}
Table~\ref{tab:detection_configs} enumerates all six configurations evaluated in this study. Configurations 2--6 share the FCOS head and are used for the controlled comparison of neck design and tuning scheme; Configuration 1 (Faster R-CNN) serves as the two-stage anchor-based reference.

\begin{table*}[t]
\centering
\caption{Summary of object detection configurations evaluated in this study. 
Detection heads, necks, and backbone adaptation regimes are described in 
Sections~\ref{sec:detection_heads}, \ref{multi-scale}, 
and~\ref{sec:backbone_adaptation}, respectively.}
\label{tab:detection_configs}
\small
\setlength{\tabcolsep}{5pt}
\renewcommand{\arraystretch}{1.15}
\begin{tabular}{@{}l l l l l r r@{}}
\toprule
\textbf{\#} & \textbf{Configuration} & \textbf{Head} & \textbf{Neck} & 
\textbf{Adaptation} & \textbf{Trainable} & \textbf{Total} \\
\midrule
1 & Faster R-CNN + LoRA               & Faster R-CNN & Pseudo            & LoRA-only   & 25.44 M  & 329.33 M \\
2 & FCOS + LoRA                        & FCOS         & Pseudo            & LoRA-only   & 15.68 M  & 319.57 M \\
3 & FCOS + LoRA + Lite ViT-Adapter     & FCOS         & Lite ViT-Adapter  & LoRA-only   & 14.38 M  & 318.27 M \\
4 & FCOS + LoRA + Full ViT-Adapter     & FCOS         & Full ViT-Adapter  & LoRA-only   & 26.99 M  & 330.87 M \\
5 & FCOS + Hybrid PEFT + Lite ViT-Adapter & FCOS      & Lite ViT-Adapter  & Hybrid PEFT & 89.17 M  & 317.48 M \\
6 & FCOS + Hybrid PEFT + Full ViT-Adapter & FCOS      & Full ViT-Adapter  & Hybrid PEFT & 101.78 M & 330.09 M \\
\bottomrule
\end{tabular}
\end{table*}

All six configurations are trained under identical hyperparameters (Table~\ref{tab:shared_hyperparameters}). The dataset is derived from a single HLS v2.0 Sentinel-2 tile (tile id: T13TGE, year: 2022) covering agricultural cropland in eastern Colorado, processed at 30 m resolution using six spectral bands (B02= Blue, B03= Green, B04= Red, B8A= NIR\_Narrow, B11= SWIR\_1, B12= SWIR\_2). Ground truth bounding boxes are generated from the 2022 USDA CDL layer via~\ref{sec:bbox_generation}. The HLS tile is partitioned spatially into 70 \% / 15 \% / 15 \% training, validation, and test zones using a fixed split seed of 42. From within each zone, a TorchGeo~\cite{stewart2025torchgeo} Random Batch GeoSampler independently draws $256x256$-pixel patches, yielding 5,000 training, 500 validation patches per epoch and all mAPs are calculated on the validation set. We train all models on a single NVIDIA A100 GPU with 16 CPU cores and 80 GB of system RAM. Table~\ref{tab:shared_hyperparameters} summarises the remaining training hyperparameters.

\begin{table}[H]
\centering
\caption{Shared training hyperparameters used across all configurations.}
\label{tab:shared_hyperparameters}
\renewcommand{\arraystretch}{1.15}
\begin{tabular}{p{3.2cm} p{4.3cm}}
\hline
\textbf{Hyperparameter} & \textbf{Value} \\
\hline
Optimizer & AdamW \\
Weight decay &  $1 \times 10^{-4}$ \\
Initial learning rate & $1 \times 10^{-4}$ \\
LR schedule & CosineAnnealingLR, $T_{\max}=200$, $\eta_{\min}=1 \times 10^{-6}$ \\
Batch size & 64 \\
Max epochs & 200 \\
Train / Val / Test split & 0.70 / 0.15 / 0.15 \\
Split seed & 42 \\
Global random seed & 2 \\
\hline
\end{tabular}
\end{table}

\subsection{Evaluation Metrics}
\label{sec:eva_metrics}
We evaluate all configurations using mean Average Precision (mAP) at IoU thresholds of 0.50 (mAP@50) and 0.75 (mAP@75) as primary metrics. For a predicted box to count as a true positive, its Intersection over Union (IoU) with the matched ground truth box must meet or exceed the  threshold. We report one extra mAP which denotes
average precision over IoU thresholds 0.50:0.05:0.95. We additionally report results under two bounding box regression loss functions: GIoU~\cite{rezatofighi2019generalized} and DIoU loss~\cite{zheng2020distance}. 
GIoU is defined as:
\begin{equation}
    \mathcal{L}_{\text{GIoU}} = 1 - \text{IoU} + 
    \frac{|\mathcal{C}| - |b \cup b^{\text{gt}}|}{|\mathcal{C}|}
    \label{eq:giou}
\end{equation}

\noindent where $b$ and $b^{\text{gt}}$ are the predicted and ground-truth 
boxes, and $\mathcal{C}$ is the smallest enclosing box of both. Distance IoU (DIoU) augments this with a center-point penalty:

\begin{equation}
    \mathcal{L}_{\text{DIoU}} = 1 - \text{IoU} + 
    \frac{\rho^{2}(b_{c},\, b^{\text{gt}}_{c})}{c^{2}}
    \label{eq:diou}
\end{equation}

\noindent where $b_{c}$ and $b^{\text{gt}}_{c}$ are the centers of the 
predicted and ground-truth boxes, $\rho(\cdot)$ is the Euclidean distance, 
and $c$ is the diagonal length of $\mathcal{C}$.

GIoU penalises the gap between predicted and ground truth boxes even when they do not overlap; DIoU additionally minimises the normalised center-point distance, which is particularly relevant for fallow parcels whose irregular shapes make center-aware regression beneficial. Both loss variants are trained and evaluated independently.

To characterise detection quality at the image level, we further evaluate the trained model on 300 held-out test patches and report per-patch F1 scores in Section~\ref{sec:qualitative_results_error}.. For each 256 × 256 patch, a predicted box is counted as a true positive if its IoU with a ground-truth box meets or exceeds 0.5; precision and recall are derived from all boxes in the patch, and F1 is their harmonic mean. Unlike mAP, which aggregates over the full validation set and multiple IoU thresholds, per-patch F1 exposes localized failure modes — missed detections and false positives — that aggregate statistics can obscure. This metric is used for error analysis only; all model comparisons and rankings are based on mAP@50 and mAP@75.

\section{Results}
\subsection{Overall Detection Performance}
 Table~\ref{tab:fcos_giou_diou_results} reports results for all six configurations trained with GIoU and DIoU box regression losses. The top-performing configuration across both loss paradigms is FCOS with Hybrid PEFT and Lite ViT-Adapter: mAP@50 = 0.9479 (DIoU-trained), a 25.70\% relative improvement over the anchor-based Faster-RCNN baseline (DIoU-trained). The GIoU-trained variant of the same configuration achieves the highest mAP@75 = 0.8643. Sections 4.2–4.4 analyse the contribution of head choice, adapter design, and PEFT schemes in turn.

\begin{table}[t]
\centering
\caption{Detection results across configurations. Bold indicates the best 
result within each loss setting.}
\label{tab:fcos_giou_diou_results}
\small
\setlength{\tabcolsep}{5pt}
\renewcommand{\arraystretch}{1.3}
\begin{tabularx}{\columnwidth}{@{}X c c c@{}}
\toprule
\textbf{Configuration} & \textbf{mAP$_{50}$} & \textbf{mAP$_{75}$} & \textbf{mAP} \\
\midrule
\multicolumn{4}{@{}l}{\textit{GIoU Loss}} \\
\midrule
Faster R-CNN + LoRA          & 0.7603 & 0.5310 & 0.4825 \\
FCOS + LoRA                  & 0.7954 & 0.6549 & 0.6119 \\
FCOS + LoRA + Lite VA        & 0.8301 & 0.6865 & 0.6430 \\
FCOS + LoRA + Full VA        & 0.9156 & 0.7973 & 0.7469 \\
FCOS + H-PEFT + Lite VA      & \textbf{0.9464} & \textbf{0.8643} & \textbf{0.8166} \\
FCOS + H-PEFT + Full VA      & 0.9277 & 0.8211 & 0.7765 \\
\midrule
\multicolumn{4}{@{}l}{\textit{DIoU Loss}} \\
\midrule
Faster R-CNN + LoRA          & 0.7541 & 0.5309 & 0.4808 \\
FCOS + LoRA                  & 0.8025 & 0.6568 & 0.6124 \\
FCOS + LoRA + Lite VA        & 0.8370 & 0.6857 & 0.6423 \\
FCOS + LoRA + Full VA        & 0.9154 & 0.7964 & 0.7501 \\
FCOS + H-PEFT + Lite VA      & \textbf{0.9479} & \textbf{0.8614} & \textbf{0.8172} \\
FCOS + H-PEFT + Full VA      & 0.9321 & 0.8228 & 0.7767 \\
\bottomrule
\end{tabularx}
\end{table}

\subsection{Effect of Detection Head}
Holding backbone (Prithvi-EO-2.0), neck (pseudo multi-scale), and adaptation scheme fixed (LoRA-only), switching from Faster R-CNN to FCOS yields a clear mAP@50 gain. Faster R-CNN achieves mAP@50 = 0.7603 (GIoU) and 0.7551 (DIoU), with mAP@75 of 0.5310 and 0.5309 respectively. Switching to FCOS improves performance across both losses and both metrics: the DIoU-trained FCOS variant reaches mAP@50 = 0.8025 (+6.42\% relative over the FR-CNN DIoU baseline) and mAP@75 = 0.6568, while the GIoU-trained variant achieves 0.7954 and 0.6549. Unlike the two-stage RPN pipeline, anchor-free centerness prediction under the same pseudo multi-scale neck produces consistently better-localised detections even without spatial priors, suggesting that FCOS's per-pixel regression is better suited to the irregular geometry of fallow parcels than proposal refinement at this neck complexity. DIoU configurations lead marginally at mAP@50 for FCOS, while GIoU leads marginally for Faster R-CNN; this loss–head interaction persists through all adapter configurations. DIoU trained FCOS + H-PEFT + Lite VA ultimately reaches mAP@75 = 0.8643, approximately 33 pp above the Faster R-CNN baseline, confirming that the FCOS heads captures substantially more localization gain from spatial prior fusion (section~\ref{sec:effect_of_adapter_design}).

\subsection{Effect of Adapter Design}
\label{sec:effect_of_adapter_design}

Within the LoRA adaptation scheme, introducing a ViT-Adapter neck progressively improves detection across all metrics. Under DIoU training, adding the Lite ViT-Adapter raises mAP@50 from 0.8025 (no adapter) to 0.8370, a 4.30\% relative gain; the Full ViT-Adapter advances this to 0.9154, a further 9.37\% over Lite and 14.07\% over no adapter. The GIoU-trained configurations follow the same progression: 0.7954 → 0.8301 → 0.9156. Gains at mAP@75 are more pronounced: no-adapter to Full under DIoU training spans nearly 14 pp (0.6568 → 0.7964), reflecting improvement in both box tightness and detection count. The Full ViT-Adapter advantage under LoRA is consistent with its design: with backbone weights largely frozen, the bidirectional injector–extractor cross-attention supplies spatial context that a frozen backbone cannot produce on its own. Lite VA's output-level fusion receives only the unmodified single-stride backbone output and cannot compensate for the absence of intermediate spatial adaptation. This ordering reverses under Hybrid PEFT, as shown in Section~\ref{sec:effect_of_tuning_scheme}.

\subsection{Effect of Tuning Scheme}
\label{sec:effect_of_tuning_scheme}

Replacing LoRA-only adaptation with Hybrid PEFT produces qualitatively different gains for Lite and Full ViT-Adapter configurations, reversing the ordering established in Section~\ref{sec:effect_of_adapter_design}. Under LoRA, Full VA led Lite VA by 7.84 pp at mAP@50 (DIoU); under Hybrid PEFT, Lite VA overtakes Full VA by 1.58 pp (0.9479 vs. 0.9321). The gain magnitudes diverge substantially. For the Lite ViT-Adapter, Hybrid PEFT adds 11.09 pp at mAP@50 (DIoU: 0.8370 → 0.9479) and 17.57 pp at mAP@75 (DIoU: 0.6857 → 0.8614), a 13.25\% relative mAP@50 improvement. For the Full ViT-Adapter, the same scheme change yields only 1.67 pp at mAP@50 (0.9154 → 0.9321) and 2.64 pp at mAP@75 (0.7964 → 0.8228), a 1.82\% relative gain. Lite VA gains approximately 7× more from Hybrid PEFT than Full VA does. The GIoU-trained configurations mirror this pattern: Lite gains 14.01\% (0.8301 → 0.9464) while Full gains only 1.32\% (0.9156 → 0.9277).

Under Hybrid PEFT, GIoU-trained Lite VA achieves the highest mAP@75 across all configurations (0.8643), marginally above its DIoU-trained counterpart (0.8614), while DIoU-trained Lite VA leads at mAP@50 (0.9479 vs. 0.9464). The mechanistic basis for this asymmetric response is discussed in Section~\ref{sec:adapter_peft_interaction}.

\subsection{Qualitative Results and Error Analysis}
\label{sec:qualitative_results_error}

\begin{figure*}[t]
    \centering
    \includegraphics[width=\textwidth]{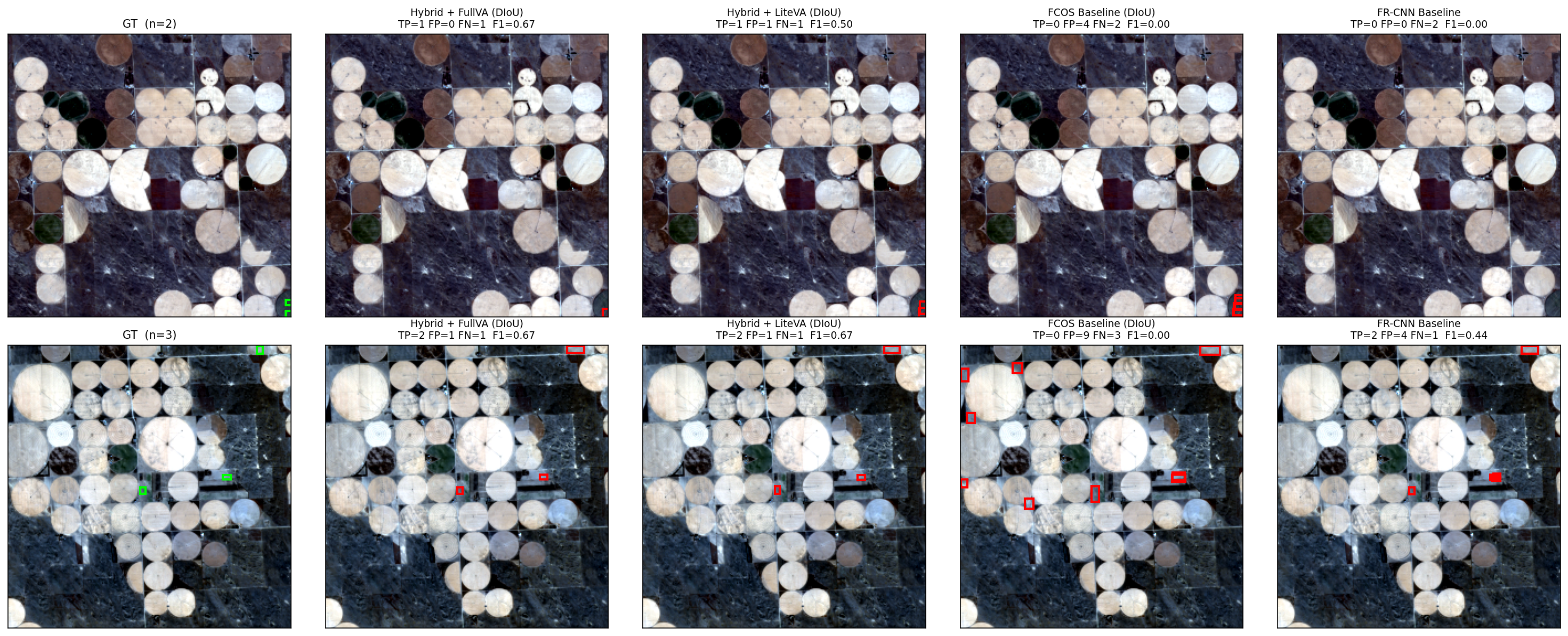}
    \caption{Qualitative detection results on representative patches from tile T13TGE. Each column shows the same $256 \times 256$ chip for ground truth and four different configurations.}
    \label{fig:comparison_panel}
\end{figure*}

Figure~\ref{fig:comparison_panel} illustrates representative detection outcomes across two patches; Table~\ref{tab:patch_f1_analysis} provides per-patch F1 statistics aggregated over 215 test patches (out of 300) containing at least one fallow bounding box. Per-patch F1 is computed at IoU $\geq$ 0.50 as defined in Section~\ref{sec:eva_metrics}; it complements mAP by revealing localized failure modes that aggregate statistics can obscure. According to per-patch F1 analysis, Hybrid+FullVA achieves the highest mean F1 of 0.939, with 82 of 215 patches detected perfectly and zero complete failures. Hybrid+LiteVA follows at 0.912, also with zero complete failures and 62 perfect patches. FR-CNN+LoRA records a mean F1 of 0.831 with one complete failure; FCOS+LoRA lags substantially at 0.549, with six patches on which it produces no detections whatsoever---every box missed. Crucially, both Hybrid PEFT configurations recover detections on all six of these hard patches, and FR-CNN recovers five of the six. The gap between FCOS+LoRA (0.549) and Hybrid+FullVA (0.939) on a per-patch basis is larger than the mAP@50 gap (0.802 vs 0.932),
indicating that the baseline's failures are concentrated rather than diffuse.

\begin{table}[H]
\centering
\caption{Per-patch F1 analysis over 215 test patches.}
\label{tab:patch_f1_analysis}
\small
\setlength{\tabcolsep}{5pt}
\renewcommand{\arraystretch}{1.3}
\begin{tabularx}{\columnwidth}{@{}X c c c@{}}
\toprule
\textbf{Configuration} & \textbf{Mean F1} & \textbf{F1$=$0} & \textbf{F1$=$1} \\
\midrule
Hybrid PEFT + Full ViT-Adapter & 0.939 & 0 & 82 \\
Hybrid PEFT + Lite ViT-Adapter & 0.912 & 0 & 62 \\
Faster R-CNN + LoRA            & 0.831 & 1 & 48 \\
FCOS + LoRA                    & 0.549 & 6 & 6  \\
\bottomrule
\end{tabularx}
\end{table}

\section{Discussions}
\subsection{System Design Implications}
The results establish FCOS + Hybrid PEFT + Lite ViT-Adapter with DIoU loss as the recommended configuration for fallow detection with Prithvi-EO-2.0 as a backbone. Three design decisions jointly account for its performance. First, anchor-free centerness prediction is better suited to the irregular, variable-size geometry of fallow fields than
RPN-based proposal generation~\cite{wen2023comprehensive}; the head choice matters most once spatial priors are available, explaining why the mAP@75 advantage Faster R-CNN holds under the adapter-free setting reverses substantially with adapter introduction. Second, DIoU regression consistently improves mAP@50 over GIoU across all configurations by
enforcing center-point alignment in addition to overlap area, which is particularly relevant for fields whose bounding boxes only approximate their true spatial extent. Third, output-level spatial prior fusion via the Lite Adapter enables cleaner interaction with partial backbone unfreezing than more complex bidirectional injection,
for reasons examined in Section~\ref{sec:adapter_peft_interaction}.

\subsection{Adapter–PEFT Interaction}
\label{sec:adapter_peft_interaction}
The most consequential finding is the asymmetric response of Lite and Full ViT-Adapters to Hybrid PEFT: Full VA's 7.84 pp LoRA-scheme advantage inverts to a 1.58 pp deficit under Hybrid PEFT, with Lite VA gaining approximately 7× more from backbone unfreezing. We interpret this as a functional substitution effect. Full VA's bidirectional injector–extractor cross-attention and Hybrid PEFT's selective backbone unfreezing address the same representational bottleneck: extracting spatially heterogeneous features from a globally-pooling ViT. Applied together, they provide redundant capacity targeting the same gap. Lite VA's output-only additive fusion does not modify internal backbone computations, making it structurally complementary to backbone unfreezing: the two mechanisms operate at different points in
the network and reinforce rather than substitute for each other. Loss curve behaviour is consistent with this interpretation. Lite VA + Hybrid PEFT converges faster in early epochs, reflecting the unlocking of a previously bottlenecked learning signal rather than destabilised optimisation.
Similar final training losses alongside divergent mAP outcomes indicate a generalisation difference rather than a
fitting difference. This interpretation is mechanistically consistent with the observed data but is not supported by formal gradient analysis, which is outside the scope of this work.
\subsection{Parameter Efficiency}
Table~\ref{tab:vram_params_map} compares peak training VRAM and trainable parameters against best mAP@50. FCOS + H-PEFT + Lite VA achieves the highest mAP@50 at the lowest VRAM among Hybrid PEFT configurations (24.3 GB), using 6.5 GB less than FCOS + H-PEFT + Full VA while outperforming it. For settings where trainable parameter count is the binding constraint, FCOS + LoRA + Full VA (27M trainable) achieves mAP@50 = 0.9154 with 62M fewer trainable parameters than Hybrid + Lite VA, sacrificing 3.25 pp at higher peak VRAM.

\begin{table}[t]
\centering
\caption{Peak training VRAM, trainable parameters, and best mAP@50; bold indicates the best overall result.}
\label{tab:vram_params_map}
\scriptsize
\setlength{\tabcolsep}{3pt}
\renewcommand{\arraystretch}{1.15}
\begin{tabularx}{\columnwidth}{@{}Xccc@{}}
\toprule
\textbf{Configuration} & \textbf{VRAM (GB)} & \textbf{Trainable (M)} & \textbf{mAP@50} \\
\midrule
Faster R-CNN + LoRA (GIoU) & 18.4 & $\approx 25$ & 0.7603 \\
FCOS + LoRA (DIoU) & 21.7 & $\approx 16$ & 0.8025 \\
FCOS + LoRA + Lite VA (DIoU) & 22.5 & $\approx 14$ & 0.8370 \\
FCOS + LoRA + Full VA (DIoU) & 29.9 & $\approx 27$ & 0.9154 \\
FCOS + H-PEFT + Lite VA (DIoU) & 24.3 & $\approx 89$ & \textbf{0.9479} \\
FCOS + H-PEFT + Full VA (DIoU) & 30.8 & $\approx 101$ & 0.9321 \\
\bottomrule
\end{tabularx}
\end{table}

\subsection{Contextual Comparison with Existing System}
The most directly comparable prior work using Prithvi-EO-2.0 for agricultural parcel detection is Hsu et al.~\cite{hsu2025geospatial}, who evaluated Prithvi-EO-1.0 on the EuroCrops dataset using a Mask R-CNN head with full end-to-end backbone fine-tuning. Their best result on EuroCrops achieves mAP@50 = 0.607. We use this as a conceptual reference point rather than a direct benchmark. Two observations from our results are informative. Our Faster R-CNN + LoRA (GIoU-trained) configuration, the closest architectural analogue to Hsu et al.~\cite{hsu2025geospatial} pipeline, achieves mAP@50 = 0.7603, a 15.3 percentage point absolute improvement over their result (25.26\% relative). Critically, our approach replaces full end-to-end backbone fine-tuning with LoRA, reducing trainable backbone parameters from ~300M to ~25M while achieving higher mAP. This aligns with the broader PEFT literature showing that selective adaptation can match or exceed full fine-tuning on in-distribution geospatial data~\cite{marti2025fine}. Our best configuration, FCOS (DIoU-trained) with Hybrid PEFT and Lite ViT-Adapter, reaches mAP@50 = 0.948, a a 34 percentage point absolute improvement (56\% relative). The comparison suggests that targeted spatial adaptation with Prithvi-EO-2.0 outperforms the full-fine-tuning approach of Hsu et al.~\cite{hsu2025geospatial}, though domain differences preclude a direct attribution.

\subsection{Implications for the Food-Water Nexus}
Fallow land is a historically ambiguous class in CDL~\cite{lark2017measuring}, limiting accuracy of regional water-use estimates tied to irrigation cessation and crop rotation monitoring. Automated detection at the HLS tile scale provides a step toward minimizing this ambiguity: bounding-box predictions identify the spatial extent of fallow patches and can inform irrigation allocation models and multi-cropping pattern assessments without manual digitization. All results derive from a single HLS tile (T13TGE, eastern Colorado, 2022 growing season). We do not assert generalizability across regions, soil types, or growing seasons; translating these findings into operational FW
nexus monitoring requires multi-tile validation and temporal consistency analysis, which we identify as priorities in section~\ref{sec:future_scope}.
\subsection{Limitations and Future Work}
\label{sec:future_scope}
Several limitations bound these conclusions. The analysis is confined to a single HLS tile, a single growing season (2022), and treats fallow detection purely as a spatial object detection problem, leaving temporal dynamics and multi-tile generalisability unaddressed. Additionally, CDL ground truth itself is less accurate in terms of fallow labelling, introducing label noise into both training and evaluation. Axis-aligned bounding boxes cannot capture the rotational extent of diagonal field boundaries, overapproximating patch area for elongated fields. The mechanistic interpretation in Section~\ref{sec:adapter_peft_interaction} derives from performance observations and loss curve behaviour on this single-tile dataset; whether the adapter–PEFT complementarity holds for other GFM architectures (SatMAE, ScaleMAE) or across regions and seasons remains
untested. Future work should prioritise: (1) multi-tile and multi-region evaluation to assess geographic transferability; (2) multi-temporal analysis across growing seasons; (3) testing the adapter–PEFT interaction on other geospatial foundation models; and (4) instance segmentation to replace bounding-box detection, reducing area estimation
error from box looseness and enabling polygon-level fallow inventories.

\section{Conclusions}
In this paper, we showcase efficient PEFT approaches and ViT-Adapters to enable using the Prithvi-EO-2.0 model to monitor fallow fields, a historically ambiguous category in USDA CDL. We efficiently build on Prithvi-EO-2.0 for fallow object detection by evaluating two PEFT schemes, three multi-scale neck designs, and two detection heads on a single HLS tile (T13TGE, eastern Colorado, 2022). The best configuration, FCOS with Hybrid PEFT and Lite ViT-Adapter under DIoU loss, achieves mAP@50 = 0.9479, a 25.70\% relative improvement over the anchor-based, adapter-free baseline. Per GIoU, the best configuration achieves mAP@50 = 0.9464. A slightly higher accuracy for DIoU shows the effectiveness of center-aware localization for irregular fallow field detection. The best configuration also achieves a high F1 value of 0.912 without any failure on test set. The central finding is that adapter design and backbone adaptation scheme interact non-trivially: Full ViT-Adapter and Hybrid PEFT are functionally substitutable and yield diminishing returns when combined, while Lite ViT-Adapter and Hybrid PEFT are complementary, producing a 7× larger gain from backbone unfreezing. For practitioners with constrained GPU budgets, FCOS + H-PEFT + Lite VA achieves the highest mAP@50 at the lowest peak VRAM among all Hybrid configurations tested. Results demonstrate a feasible workflow for detecting an agricultural land-use type with significant impact on food and water security.

\section*{Generative AI Usage Disclosure}
During the preparation of this work, generative AI tools (e.g. Google Gemini) are used for language editing, grammar improvement, and formatting assistance. These tools are not used to generate experimental results, create datasets, or perform quantitative analysis. All AI-assisted text and formatting suggestions are reviewed and revised by the authors to ensure technical accuracy, originality, and consistency with the paper's contributions.

\bibliographystyle{ACM-Reference-Format}
\bibliography{references}

@inproceedings{ren2015fasterrcnn,
  author    = {Ren, Shaoqing and He, Kaiming and Girshick, Ross and Sun, Jian},
  title     = {Faster {R-CNN}: Towards Real-Time Object Detection with Region Proposal Networks},
  booktitle = {Advances in Neural Information Processing Systems},
  volume    = {28},
  pages     = {91--99},
  year      = {2015}
}

@inproceedings{cao2021experimental,
  title={Experimental study on the effect of loss function on object detection},
  author={Cao, Qianyu},
  booktitle={Proceedings of the 2021 International Conference on Pattern Recognition and Intelligent Systems},
  pages={81--87},
  year={2021}
}

@inproceedings{xiaofan2019introduce,
  title={Introduce GIoU into RFB net to optimize object detection bounding box},
  author={XiaoFan, Li and HaiBo, Pu and Yi, Wei and JiangChuan, Liu and HongXiang, Xu},
  booktitle={Proceedings of the 5th International Conference on Communication and Information Processing},
  pages={108--113},
  year={2019}
}

@misc{gao2026introduction,
  title={Introduction to the Special Issue on GeoAI Foundation Models and Their Applications, Part I},
  author={Gao, Song and Lunga, Dalton and Yang, Lexie and Newsam, Shawn and Martins, Bruno},
  journal={ACM Transactions on Spatial Algorithms and Systems},
  volume={12},
  number={3},
  pages={1--4},
  year={2026},
  publisher={ACM New York, NY}
}

@inproceedings{nguyen2025foundation,
  title={Foundation Models for Semantic Segmentation of Thick/Thin Clouds and Cloud-shadows: A Comparative Study},
  author={Nguyen, Calvin and Vatsavai, Ranga Raju},
  booktitle={Proceedings of the 33rd ACM International Conference on Advances in Geographic Information Systems},
  pages={549--552},
  year={2025}
}

@article{white1970fallowing,
  title={Fallowing, crop rotation, and crop yields in Roman times},
  author={White, Kenneth D},
  journal={Agricultural History},
  volume={44},
  number={3},
  pages={281--290},
  year={1970},
  publisher={JSTOR}
}

@article{schneider2023eurocrops,
  title={EuroCrops: The largest harmonized open crop dataset across the European Union},
  author={Schneider, Maja and Schelte, Tobias and Schmitz, Felix and K{\"o}rner, Marco},
  journal={Scientific Data},
  volume={10},
  number={1},
  pages={612},
  year={2023},
  publisher={Nature Publishing Group UK London}
}

@article{zarczynski2023role,
  title={The role of fallows in sustainable development},
  author={{\.Z}arczy{\'n}ski, Piotr Jaros{\l}aw and Krzebietke, S{\l}awomir J{\'o}zef and Sienkiewicz, Stanis{\l}aw and Wierzbowska, Jadwiga},
  journal={Agriculture},
  volume={13},
  number={12},
  pages={2174},
  year={2023},
  publisher={MDPI}
}

@article{albrecht2018water,
  title={The Water-Energy-Food Nexus: A systematic review of methods for nexus assessment},
  author={Albrecht, Tamee R and Crootof, Arica and Scott, Christopher A},
  journal={Environmental Research Letters},
  volume={13},
  number={4},
  pages={043002},
  year={2018},
  publisher={IOP Publishing}
}

@article{zeleke2017fallow,
  title={Fallow management increases soil water and nitrogen storage},
  author={Zeleke, Ketema Tilahun},
  journal={Agricultural Water Management},
  volume={186},
  pages={12--20},
  year={2017},
  publisher={Elsevier}
}

@article{wen2023comprehensive,
  title={A comprehensive survey of oriented object detection in remote sensing images},
  author={Wen, Long and Cheng, Yu and Fang, Yi and Li, Xinyu},
  journal={Expert Systems with Applications},
  volume={224},
  pages={119960},
  year={2023},
  publisher={Elsevier}
}

@article{szwarcman2025prithvieo2,
  author  = {Szwarcman, Daniela and Roy, Sujit and Fraccaro, Paolo and Gislason, Orsteinn Eli and Blumenstiel, Benedikt and Ghosal, Rinki and De Oliveira, Pedro Henrique and Almeida, Joao Lucas de Sousa and Sedona, Rocco and Kang, Yanghui and Chakraborty, Srija and Wang, Sizhe and Gomes, Carlos and Kumar, Ankur and Gaur, Vishal and Truong, Myscon and Godwin, Denys and Khallaghi, Sam and Lee, Hyunho and Hsu, Chia Yu and Asanjan, Ata Akbari and Mujeci, Besart and Shidham, Disha and Balogun, Rufai Omowunmi and Kolluru, Venkatesh and Keenan, Trevor and Arevalo, Paulo and Li, Wenwen and Alemohammad, Hamed and Olofsson, Pontus and Mayer, Timothy and Hain, Christopher and Kennedy, Robert and Zadrozny, Bianca and Bell, David and Cavallaro, Gabriele and Watson, Campbell and Maskey, Manil and Ramachandran, Rahul and Moreno, Juan Bernabe},
  title   = {{Prithvi-EO-2.0}: A Versatile Multi-Temporal Foundation Model for Earth Observation Applications},
  journal = {IEEE Transactions on Geoscience and Remote Sensing},
  year    = {2025},
  doi     = {10.1109/TGRS.2025.3642610}
}

@inproceedings{lin2017feature,
  title={Feature pyramid networks for object detection},
  author={Lin, Tsung-Yi and Doll{\'a}r, Piotr and Girshick, Ross and He, Kaiming and Hariharan, Bharath and Belongie, Serge},
  booktitle={Proceedings of the IEEE conference on computer vision and pattern recognition},
  pages={2117--2125},
  year={2017}
}

@article{chen2023mapping,
  title={Mapping center pivot irrigation systems in global arid regions using instance segmentation and analyzing their spatial relationship with freshwater resources},
  author={Chen, Fen and Zhao, Haojie and Roberts, Dar and Van de Voorde, Tim and Batelaan, Okke and Fan, Tao and Xu, Wenbo},
  journal={Remote Sensing of Environment},
  volume={297},
  pages={113760},
  year={2023},
  publisher={Elsevier}
}

@article{zheng2026comprehensive,
  title={A comprehensive review of agricultural parcel and boundary delineation from remote sensing images: Recent progress and future perspectives},
  author={Zheng, Juepeng and Ye, Zi and Wen, Yibin and Huang, Jianxi and Zhang, Zhiwei and Li, Qingmei and Hu, Qiong and Xu, Baodong and Zhao, Lingyuan and Fu, Haohuan},
  journal={IEEE Geoscience and Remote Sensing Magazine},
  year={2026},
  publisher={IEEE}
}

@article{upcott2023new,
  title={A new approach to characterising and predicting crop rotations using national-scale annual crop maps},
  author={Upcott, Emily V and Henrys, Peter A and Redhead, John W and Jarvis, Susan G and Pywell, Richard F},
  journal={Science of the Total Environment},
  volume={860},
  pages={160471},
  year={2023},
  publisher={Elsevier}
}

@book{greb1979reducing,
  title={Reducing drought effects on croplands in the west-central Great Plains},
  author={Greb, BW},
  number={420},
  year={1979},
  publisher={US Department of Agriculture, Science and Education Administration}
}

@article{anderson1999alternative,
  title={Alternative crop rotations for the central Great Plains},
  author={Anderson, RL and Bowman, RA and Nielsen, DC and Vigil, MF and Aiken, RM and Benjamin, JG},
  journal={Journal of Production Agriculture},
  volume={12},
  number={1},
  pages={95--99},
  year={1999},
  publisher={Wiley Online Library}
}

@article{jiang2024trade,
  title={Trade-off between the future water resource utilization and grain production in a water-deficient region from the perspective of the Water- Land- Grain nexus},
  author={Jiang, Luguang and Liu, Ye and Yang, Cheng},
  journal={Journal of Hydrology},
  volume={640},
  pages={131697},
  year={2024},
  publisher={Elsevier}
}

@article{jain2023water,
  title={Water-energy-food-ecosystem nexus in India—A review of relevant studies, policies, and programmes},
  author={Jain, Sharad K and Sikka, Alok K and Alam, Mohammad Faiz},
  journal={Frontiers in Water},
  volume={5},
  pages={1128198},
  year={2023},
  publisher={Frontiers Media SA}
}

@article{hung2022investigating,
  title={Investigating uncertainties in human adaptation and their impacts on water scarcity in the Colorado river Basin, United States},
  author={Hung, Fengwei and Son, Kyongho and Yang, YC Ethan},
  journal={Journal of hydrology},
  volume={612},
  pages={128015},
  year={2022},
  publisher={Elsevier}
}

@article{peng2022domain,
  title={Domain adaptation in remote sensing image classification: A survey},
  author={Peng, Jiangtao and Huang, Yi and Sun, Weiwei and Chen, Na and Ning, Yujie and Du, Qian},
  journal={IEEE Journal of Selected Topics in Applied Earth Observations and Remote Sensing},
  volume={15},
  pages={9842--9859},
  year={2022},
  publisher={IEEE}
}

@article{belgiu2018sentinel,
  title={Sentinel-2 cropland mapping using pixel-based and object-based time-weighted dynamic time warping analysis},
  author={Belgiu, Mariana and Csillik, Ovidiu},
  journal={Remote sensing of environment},
  volume={204},
  pages={509--523},
  year={2018},
  publisher={Elsevier}
}

@article{begue2018remote,
  title={Remote sensing and cropping practices: A review},
  author={B{\'e}gu{\'e}, Agn{\`e}s and Arvor, Damien and Bellon, Beatriz and Betbeder, Julie and De Abelleyra, Diego and PD Ferraz, Rodrigo and Lebourgeois, Valentine and Lelong, Camille and Sim{\~o}es, Margareth and R. Ver{\'o}n, Santiago},
  journal={Remote Sensing},
  volume={10},
  number={1},
  pages={99},
  year={2018},
  publisher={MDPI}
}

@article{subedi2022drivers,
  title={Drivers and consequences of agricultural land abandonment and its reutilisation pathways: A systematic review},
  author={Subedi, Yuba Raj and Kristiansen, Paul and Cacho, Oscar},
  journal={Environmental Development},
  volume={42},
  pages={100681},
  year={2022},
  publisher={Elsevier}
}

@article{oliphant2024automated,
  title={Automated Cropland Fallow Algorithm (ACFA) for the Northern Great Plains of USA},
  author={Oliphant, Adam J and Thenkabail, Prasad S and Teluguntla, Pardhasaradhi G and Aneece, Itiya P and Foley, Daniel J and McCormick, Richard L},
  journal={International Journal of Digital Earth},
  volume={17},
  number={1},
  pages={2337221},
  year={2024},
  publisher={Taylor \& Francis}
}

@article{song2023cropland,
  title={Cropland fallow reduces agricultural water consumption by 303 million tons annually in Gansu Province, China},
  author={Song, Wen and Song, Wei},
  journal={Science of the Total Environment},
  volume={879},
  pages={163013},
  year={2023},
  publisher={Elsevier}
}

@article{siebet2010global,
author = {Siebert, Stefan and Portmann, Felix and Doell, Petra},
year = {2010},
month = {06},
pages = {1625-1643},
title = {Global Patterns of Cropland Use Intensity},
volume = {2},
journal = {Remote Sensing},
doi = {10.3390/rs2071625}
}

@article{daskalova2023abandoning,
  title={Abandoning land transforms biodiversity},
  author={Daskalova, Gergana N and Kamp, Johannes},
  journal={Science},
  volume={380},
  number={6645},
  pages={581--583},
  year={2023},
  publisher={American Association for the Advancement of Science}
}

@article{lark2017measuring,
  title={Measuring land-use and land-cover change using the US department of agriculture’s cropland data layer: Cautions and recommendations},
  author={Lark, Tyler J and Mueller, Richard M and Johnson, David M and Gibbs, Holly K},
  journal={International journal of applied earth observation and geoinformation},
  volume={62},
  pages={224--235},
  year={2017},
  publisher={Elsevier}
}

@article{kozak2021impact,
  title={Impact assessment of the long-term fallowed land on agricultural soils and the possibility of their return to agriculture},
  author={Kozak, Ma{\l}gorzata and Pude{\l}ko, Rafa{\l}},
  journal={Agriculture},
  volume={11},
  number={2},
  pages={148},
  year={2021},
  publisher={MDPI}
}

@article{salamon2011plant,
  title={Plant species effects on soil macrofauna density in grassy arable fallows of different age},
  author={Salamon, J{\"o}rg-Alfred and Wissuwa, Janet and Jagos, Stephan and Koblm{\"u}ller, Monika and Ozinger, Oxana and Winkler, Christine and Frank, Thomas},
  journal={European Journal of Soil Biology},
  volume={47},
  number={2},
  pages={129--137},
  year={2011},
  publisher={Elsevier}
}

@article{yin2020monitoring,
  title={Monitoring cropland abandonment with Landsat time series},
  author={Yin, He and Brand{\~a}o Jr, Amintas and Buchner, Johanna and Helmers, David and Iuliano, Benjamin G and Kimambo, Niwaeli E and Lewi{\'n}ska, Katarzyna E and Razenkova, Elena and Rizayeva, Afag and Rogova, Natalia and others},
  journal={Remote Sensing of Environment},
  volume={246},
  pages={111873},
  year={2020},
  publisher={Elsevier}
}

@article{foerster2012crop,
  title={Crop type mapping using spectral--temporal profiles and phenological information},
  author={Foerster, Saskia and Kaden, Klaus and Foerster, Michael and Itzerott, Sibylle},
  journal={Computers and Electronics in Agriculture},
  volume={89},
  pages={30--40},
  year={2012},
  publisher={Elsevier}
}

@article{tong2020forgotten,
  title={The forgotten land use class: Mapping of fallow fields across the Sahel using Sentinel-2},
  author={Tong, Xiaoye and Brandt, Martin and Hiernaux, Pierre and Herrmann, Stefanie and Rasmussen, Laura Vang and Rasmussen, Kjeld and Tian, Feng and Tagesson, Torbern and Zhang, Wenmin and Fensholt, Rasmus},
  journal={Remote Sensing of Environment},
  volume={239},
  pages={111598},
  year={2020},
  publisher={Elsevier}

}

@article{boryan2011monitoring,
  title={Monitoring US agriculture: the US department of agriculture, national agricultural statistics service, cropland data layer program},
  author={Boryan, Claire and Yang, Zhengwei and Mueller, Rick and Craig, Mike},
  journal={Geocarto International},
  volume={26},
  number={5},
  pages={341--358},
  year={2011},
  publisher={Taylor \& Francis}
}

@article{ma2026harvesting,
  title={Harvesting AlphaEarth: Benchmarking the geospatial foundation model for agricultural downstream tasks},
  author={Ma, Yuchi and Shen, Yawen and Swatantran, Anu and Lobell, David B},
  journal={International Journal of Applied Earth Observation and Geoinformation},
  volume={149},
  pages={105258},
  year={2026},
  publisher={Elsevier}
}

@inproceedings{vatsavai2024geospatial,
  title={Geospatial foundation models: Recent advances and applications},
  author={Vatsavai, Ranga Raju},
  booktitle={Proceedings of the 12th ACM SIGSPATIAL International Workshop on Analytics for Big Geospatial Data},
  pages={30--33},
  year={2024}
}

@article{hong2024spectralgpt,
  title={SpectralGPT: Spectral remote sensing foundation model},
  author={Hong, Danfeng and Zhang, Bing and Li, Xuyang and Li, Yuxuan and Li, Chenyu and Yao, Jing and Yokoya, Naoto and Li, Hao and Ghamisi, Pedram and Jia, Xiuping and others},
  journal={IEEE transactions on pattern analysis and machine intelligence},
  volume={46},
  number={8},
  pages={5227--5244},
  year={2024},
  publisher={IEEE}
}

@article{jakubik2023foundation,
  title={Foundation models for generalist geospatial artificial intelligence},
  author={Jakubik, Johannes and Roy, Sujit and Phillips, CE and Fraccaro, Paolo and Godwin, Denys and Zadrozny, Bianca and Szwarcman, Daniela and Gomes, Carlos and Nyirjesy, Gabby and Edwards, Blair and others},
  journal={arXiv preprint arXiv:2310.18660},
  year={2023}
}

@article{hsu2025geospatial,
  title={Geospatial foundation models for image analysis: Evaluating and enhancing NASA-IBM Prithvi’s domain adaptability},
  author={Hsu, Chia-Yu and Li, Wenwen and Wang, Sizhe},
  journal={International Journal of Geographical Information Science},
  volume={39},
  number={9},
  pages={2096--2125},
  year={2025},
  publisher={Taylor \& Francis}
}

@inproceedings{marti2025fine,
  title={Fine-tune smarter, not harder: Parameter-efficient fine-tuning for geospatial foundation models},
  author={Marti Escofet, Francesc and Blumenstiel, Benedikt and Scheibenreif, Linus and Fraccaro, Paolo and Schindler, Konrad},
  booktitle={Joint European Conference on Machine Learning and Knowledge Discovery in Databases},
  pages={516--532},
  year={2025},
  organization={Springer}
}

@article{hu2022lora,
  title={Lora: Low-rank adaptation of large language models.},
  author={Hu, Edward J and Shen, Yelong and Wallis, Phillip and Allen-Zhu, Zeyuan and Li, Yuanzhi and Wang, Shean and Wang, Liang and Chen, Weizhu and others},
  journal={Iclr},
  volume={1},
  number={2},
  pages={3},
  year={2022}
}

@inproceedings{tian2019fcos,
  title={Fcos: Fully convolutional one-stage object detection},
  author={Tian, Zhi and Shen, Chunhua and Chen, Hao and He, Tong},
  booktitle={Proceedings of the IEEE/CVF international conference on computer vision},
  pages={9627--9636},
  year={2019}
}

@article{song2022mapping,
  title={Mapping the spatial and temporal patterns of fallow land in mountainous regions of China},
  author={Song, Wen and Prishchepov, Alexander V and Song, Wei},
  journal={International Journal of Digital Earth},
  volume={15},
  number={1},
  pages={2148--2167},
  year={2022},
  publisher={Taylor \& Francis}
}

@article{stewart2025torchgeo,
  title={Torchgeo: deep learning with geospatial data},
  author={Stewart, Adam J and Robinson, Caleb and Corley, Isaac A and Ortiz, Anthony and Lavista Ferres, Juan M and Banerjee, Arindam},
  journal={ACM Transactions on Spatial Algorithms and Systems},
  volume={11},
  number={4},
  pages={1--28},
  year={2025},
  publisher={ACM New York, NY}
}

@inproceedings{rezatofighi2019generalized,
  title={Generalized intersection over union: A metric and a loss for bounding box regression},
  author={Rezatofighi, Hamid and Tsoi, Nathan and Gwak, JunYoung and Sadeghian, Amir and Reid, Ian and Savarese, Silvio},
  booktitle={Proceedings of the IEEE/CVF conference on computer vision and pattern recognition},
  pages={658--666},
  year={2019}
}

@inproceedings{zheng2020distance,
  title={Distance-IoU loss: Faster and better learning for bounding box regression},
  author={Zheng, Zhaohui and Wang, Ping and Liu, Wei and Li, Jinze and Ye, Rongguang and Ren, Dongwei},
  booktitle={Proceedings of the AAAI conference on artificial intelligence},
  volume={34},
  number={07},
  pages={12993--13000},
  year={2020}
}

@article{chen2022vision,
  title={Vision transformer adapter for dense predictions},
  author={Chen, Zhe and Duan, Yuchen and Wang, Wenhai and He, Junjun and Lu, Tong and Dai, Jifeng and Qiao, Yu},
  journal={arXiv preprint arXiv:2205.08534},
  year={2022}
}

\end{document}